 \documentclass[pmlr,twocolumn,10pt]{jmlr} % W&CP article

\usepackage{booktabs}
\usepackage[load-configurations=version-1]{siunitx} % newer version 

\usepackage[utf8]{inputenc} % allow utf-8 input
\usepackage[T1]{fontenc}    % use 8-bit T1 fonts
\usepackage{hyperref}       % hyperlinks
\usepackage{booktabs}       % professional-quality tables
\usepackage{amsfonts}       % blackboard math symbols
\usepackage{nicefrac}       % compact symbols for 1/2, etc.
\usepackage{microtype}      % microtypography
\usepackage{xcolor}         % colors
\usepackage{soul}
\usepackage{mdframed}
\usepackage[export]{adjustbox}

\theorembodyfont{\upshape}
\theoremheaderfont{\scshape}
\theorempostheader{:}
\theoremsep{\newline}

\jmlrpages{}
\jmlrworkshop{Machine Learning for Health (ML4H) 2021} % W&CP title

\newcommand\indepm{\protect\mathpalette{\protect\independenT}{\perp}}
\def\independenT#1#2{\mathrel{\rlap{$#1#2$}\mkern2mu{#1#2}}}
\DeclareMathOperator*{\indep}{\indepm}

\title{ADCB: An Alzheimer's disease benchmark for evaluating observational estimators of causal effects}

\author{%
\Name{Newton Mwai Kinyanjui} \Email{mwai@chalmers.se}\\
\addr Chalmers University of Technology, Sweden
\AND
\Name{Fredrik D. Johansson} \Email{fredrik.johansson@chalmers.se}\\
\addr Chalmers University of Technology, Sweden
}

\begin{document}

\maketitle

\begin{abstract}
Simulators make unique benchmarks for causal effect estimation since they do not rely on unverifiable assumptions or the ability to intervene on real-world systems, but are often too simple to capture important aspects of real applications.
We propose a simulator of Alzheimer's disease aimed at modeling intricacies of healthcare data while enabling benchmarking of causal effect and policy estimators. We fit the system to the Alzheimer's Disease Neuroimaging Initiative (ADNI) dataset and ground hand-crafted components in results from comparative treatment trials and observational treatment patterns. The simulator includes parameters which alter the nature and difficulty of the causal inference tasks, such as latent variables, effect heterogeneity, length of observed history, behavior policy and sample size. We use the simulator to compare estimators of average and conditional treatment effects.
\end{abstract}

\begin{keywords}
Causal effects; benchmark
\end{keywords}

\section{Introduction}
\label{sec:intro}
Evaluating learned decision-making policies and observational estimates of causal effects is challenging. Real-world implementation is often not an option and basing evaluation on observational data must rely on strong assumptions and access to large samples~\citep{rosenbaum2010design}. As a result, the research community often turns to simulators for benchmarking~\citep{dorie2019automated,chan2021medkit}.

Simulated data have many advantages but often lack the intricacies of the real thing. For example, two of the most widely used benchmarks in the community studying causal effects, IHDP~\citep{hill2011bayesian} and the ACIC challenge~\citep{dorie2019automated}, have response surfaces which are hand-crafted from simple building blocks without connection to real data. 
To remedy this, many simulators make use of either a) actual samples for a subset of variables, b) functions fit to real samples, or both~\citep{neal2020realcause} applies both methods.
A downside of simulating only a subset of variables is that the sample size must be fixed. See Appendix~\ref{app:related} for a more complete survey.

We propose a new benchmark for evaluating estimators of causal effects, the Alzheimer's Disease Causal estimation Benchmark (ADCB). ADCB combines the strengths of data-driven simulators with those of hand-crafted components by fitting a longitudinal causal model of patient variables to real data, while providing tunable parameters which change properties of the system and the difficulty of the benchmark. i) Causal effects are based on published results from randomized controlled trials with heterogeneity introduced through a latent variable. ii) Overlap and variance in treatment choice is controlled by different behavior policies. iii) The length of observed history is set by the user. We use the benchmark to test estimators of causal effects a) where a single time point is used to estimate average and personalized treatment effects, and b) where a time series of patient history is used.

\paragraph{Estimation of causal effects} 
The central quantity of interest in this work is the causal effect of an action $A=a$, $\Delta(a) = Y(a) - Y(0)$, defined as the difference between the potential outcomes~\citep{rubin2005causal} of $a$ and a baseline action $A=0$. We denote the set of $k$ candidate actions $\mathcal{A} = \{0, ..., k-1\}$. As $\Delta$ itself is rarely identifiable, it is common to represent the value of $a$ using the \emph{average treatment effect} (ATE), $\tau(a) = \mathbb{E}[\Delta(a)]$ or the \emph{conditional average treatment effect} (CATE) in a context $X=x$, $\tau(a \mid x) = \mathbb{E}[\Delta(a) \mid X=x]$. ATE and CATE measure how well action $a$ performs in general and in a stratum $x$, respectively. 

We study estimators of effects from samples of contexts $X$, actions $A$ and outcomes $Y$. ATE and CATE may be estimated from such data provided an adjustment set $C \subset X$ under the conditions of \emph{consistency}, $Y=Y(A)$, \emph{overlap}, $p(A=a \mid C=c) > 0$, and \emph{exchangeability}, $Y(a) \indep A \mid C$, for all $a$ and $c$.

%
% Data
%
\section{The ADCB simulator}\label{sec:data}
\label{sec:ADCB}
\label{sec:methods}

Alzheimer's disease (AD) provides an interesting test bed as the progressive nature of the condition allows for long-term study of subjects. There is evidence that AD is in fact composed of multiple disease subtypes, possibly associated with heterogeneous responses to treatment. Finally, while many treatments for AD have been developed, their effects are considered symptomatic and not affecting the underlying disease itself. This allows for simplifying structural assumptions in the simulator. Simulated treatments are drugs from the AD literature~\citep{grossberg2019present} including three cholinesterase inhibitors (ChEIs) and the N-methyl-D-aspartate receptor antagonist memantine.

\subsection{Longitudinal subject data}

We make use of %publicly available longitudinal data, of both AD patients and cognitively normal controls,
data from the Alzheimer's Disease Neuroimaging Initiative (ADNI) database (\url{http://adni.loni.usc.edu}). ADNI collects clinical data, neuroimaging data, genetic data, biological markers, and clinical and neuropsychological assessments 
%from participants at different sites in the USA and Canada 
to study cognitive impairment and AD. 
%Since its inception in 2003, several releases have been made; 
The cohorts used in this work were assembled from ADNI 1,2,3 and GO. We use trajectories of 870 unique patients where measurements are taken in 12 month intervals.

\paragraph{Patient covariates \& disease outcomes}
Subjects are represented by demographics (sex, age, education level) and various biomarkers (A$\beta$ plaques, Tau, APOE, volume of brain regions, lipids and proteins) whose detailed descriptions are provided in appendix~\ref{apd:covariates}. The specific variables used to model the time-varying context $X_t$ in this work are presented in Figure~\ref{fig:causal_graph}. The severity of Alzheimer's disease is primarily assessed based on cognitive function using tests such as the Alzheimer Disease Assessment Scale (ADAS)~\citep{rosen1984new}. We use the ADAS13 variant as our base outcome $Y_t(0)$, as it has been found to be more responsive to disease progression~\citep{cho2021disease}. ADAS13 scores take values between 0-85 where higher scores indicate worse cognitive function. The ADNI data also also contains clinical diagnosis states $D_t \in$ \{cognitively normal (CN), mild cognitive impairment (MCI),  Alzheimer’s disease (AD) \}.

\subsection{Model}
\begin{figure}[t!]%
    \centering
 
    \includegraphics[width=0.5\textwidth]{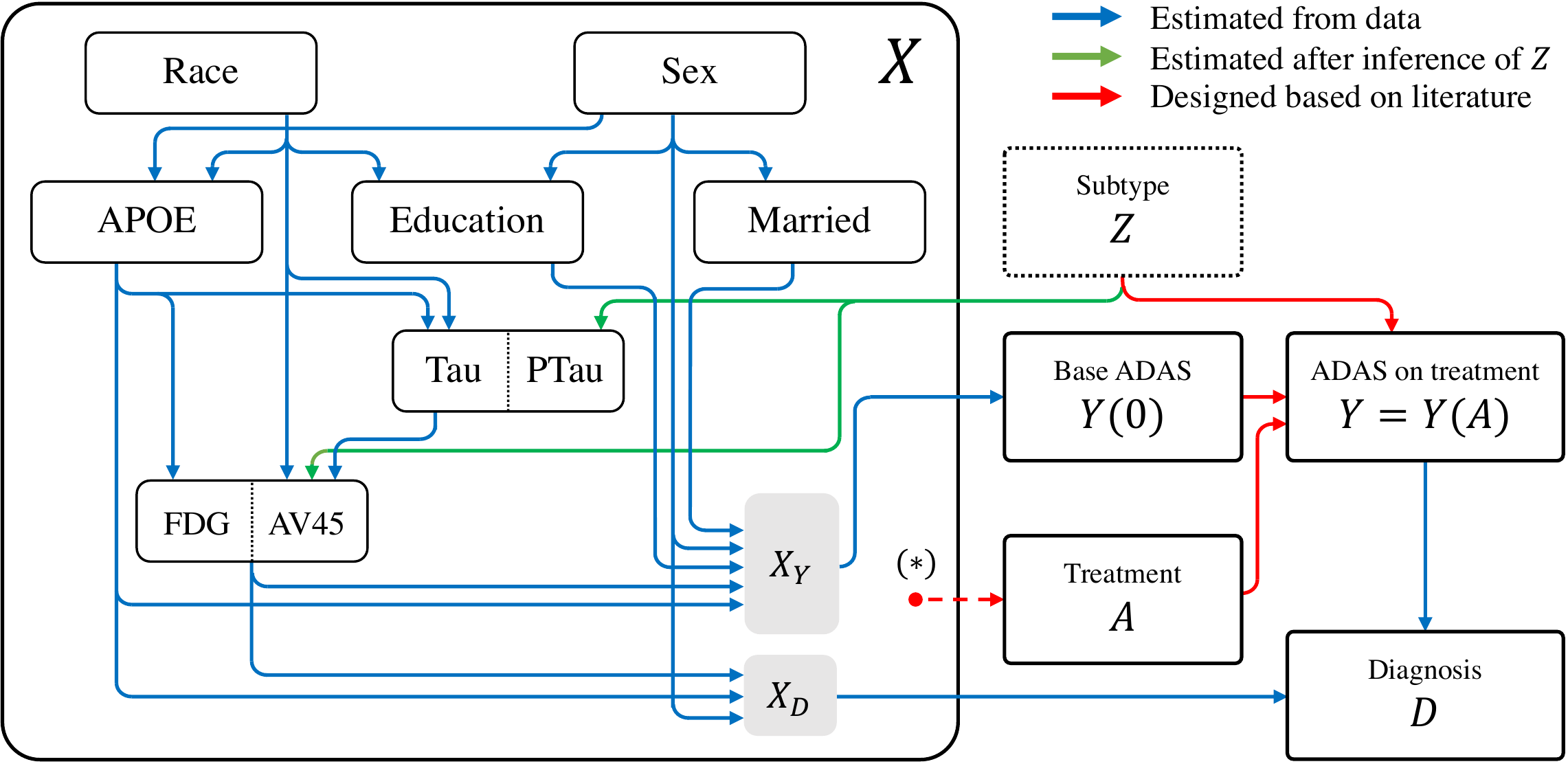}
    \caption{\label{fig:causal_graph} Assumed causal graph for a single time point. Arrows indicate causal dependencies, with color representing how the mechanism was determined. Blue dependencies were completely estimated from data, green were fit once the subtype $Z$ was inferred, and red were designed based on the AD literature. %The collections $X_Y$ and $X_D$ reduce visual clutter. %
    %Temporal dependences are shown in the Appendix. 
    }%
\end{figure}

We start by positing a causal graph for the covariates of interest at a single time point. The graph, illustrated in Figure~\ref{fig:causal_graph}, was inspired by the structure inferred from data in ~\citep{sood2020realistic} and further verified by a domain expert in Alzheimer's disease. The longitudinal model is formed by first repeating each variable, except the disease subtype $Z$, at each time step $t=1, 2, ..., T$, maintaining the causal structure of the single-time graph, see Figure~\ref{fig:causal_graph_time} in the appendix. Then, each variable is connected to the previous instance of itself;  $\mbox{Tau}_t$ is assumed to be a direct cause of $\mbox{Tau}_{t+1}$, and so on. We describe modeling choices not directly driven by the ADNI data next. 

%
% Disease subtype
%
\subsubsection{Disease subtype (latent state $Z$)}
It is believed that there are multiple subtypes of Alzheimer's disease. One of the signs of this is that in subjects, the level of so-called Amyloid-$\beta$ (A$\beta$) plaques form a clearly bimodal distribution, on the ratio of ($\frac{A\beta-42}{A\beta-40}$),~see Figure~\ref{fig:abeta} in appendix. We posit that there are two types of subjects, as indicated by a binary variables $Z\in \{0,1\}$, which, among other things, give rise to the two modes in the A$\beta$-ratio. To this end, we infer the subtype $Z$ by fitting a Gaussian mixture model with 2 components as in~\citep{dansson2021predicting}~for the A$\beta$ ratio observations of patients at baseline. We assume that $Z$ is stationary and use the model to label all observed trajectories.

\subsubsection{Treatment assignment $A$}
ADNI does not include significant data on treatments, which obstructs our vision of this dataset as a benchmark for causal effects estimators. To overcome this challenge, we designed synthetic policies for treatment assignment and treatment effects based on i) surveys of common treatments and ii) randomized controlled trials (RCT) of their effect. We begin with the former. 

Currently, there is no cure for Alzheimer's disease. However, existing drugs have been shown to have at least symptomatic effects~\citep{livingston2017lancet, farlow2008treatment}. In this work, we model a range of such drugs $a = 1, ..., 7$, for which RCT results on treatment effects are available: Donepezil 5mg,  Donepezil 10mg, Galantamine 24mg, Galantamine 32mg, Rivastigmine 12mg, Memantine 20mg,  Memantine+ChEI, see ~\citep{grossberg2019present} for an overview. We assume that the no-treatment option, $a=0$, corresponds to observations in ADNI. We simulate treatments from two simple policies $\mu_B$: %\cite{hernandez2010pharmacological}
\paragraph{Random Policy} The random policy selects each action  randomly with equal probability at each time. %Although not realistic, it corresponds to a sequentially randomized experiment.

\paragraph{Covariate-based policy} With this policy, treatments are assigned based on the diagnosis observed at the previous time step. We group treatments into 3 classes. 
Patients with mild diagnosis are assigned a randomly chosen treatment from the class with smallest ATE, those with moderate from the class with moderate ATE, and those with the most severe diagnosis from the class with the largest effect.

\begin{mdframed}[innerbottommargin=.8em,innertopmargin=0em]%
\paragraph{Overlap strength $\epsilon$}~ A tuning parameter $\epsilon \in [0, 1]$ interpolates between the random policy ($\epsilon = 1$) and the covariate-based policy $(\epsilon = 0)$ by assigning a random action with probability $\epsilon$.
\end{mdframed}

%
% Treatment assignment
%
\subsubsection{Treatment effects $\Delta$}
Consistent with the AD literature, we assume that the effects of each drug $a$ are primarily symptomatic and temporary, attenuating when treatment is stopped~\citep{grossberg2019present}. In addition, we assume that the effect is stationary in time. To this end, we endow each treatment $a$ with an additive effect $\Delta(a, Z)$, depending on the disease subtype $Z$, and posit that the cognitive function when on drug $a$ is given by $Y_t(a) = \Delta(a, Z) + Y_t(0) + \epsilon_t$. $Y_t(0)$ is estimated from observations of the ADAS13 score.

To ground our model in domain knowledge, we design $\Delta(a, Z)$ such that the average effect $\tau(a) = \mathbb{E}[\Delta(a, Z)]$ is consistent with real-world effects on cognitive function (in the ADAS-Cog scale) estimated in RCTs~\citep{grossberg2019present}. For a list of the ATEs $\tau(a)$, for $a=1, ..., k$, taken from the literature, see Appendix~\ref{apd:grossberg}. 

Given the ATE $\tau(a)$ for a treatment $a$, heterogeneity is introduced through the subtype $z\in Z$. When $Z$ is binary, as in this abstract, we let each subtype-action pair $(a, z)$ have {\sc high} or {\sc low} effect, with multiplicative margin $\gamma$, such that the opposite subtype $(a, 1-z)$ has the opposite effect for the same action.
$$
\Delta(a, z) = \left\{
\begin{array}{ll}
\frac{\tau(a)}{p(Z=z) + p(Z\neq z)\gamma}, &  \mbox{ if } \Delta(a,z) \mbox{ \sc low } \vspace{.5em} \\
\frac{\gamma\tau(a)}{p(Z=z)\gamma + p(Z \neq z)}, & \mbox{ if } \Delta(a,z) \mbox{ \sc high }\\
\end{array}
\right.~.
$$
Whether $\Delta(a,z)$ is {\sc high} or {\sc low} is determined by a look-up table designed by the user. 

\begin{mdframed}[innerbottommargin=.8em,innertopmargin=0em]%
\paragraph{Treatment effect heterogeneity $\gamma$.} The parameter $\gamma \geq 1$ controls heterogeneity in effect such that $\Delta(a, z) = \gamma \Delta(a, 1-z)$ if $\Delta(a, z)$ is {\sc high} and vice versa. $\gamma$ varies heterogeneity without changing the average treatment effect $\tau(a)$. $\gamma=1$ results in no heterogeneity.
\end{mdframed}

\subsection{Fitting the model and simulating data}

Based on the causal the graph presented in Figure~\ref{fig:causal_graph}, we learn a joint distribution of the full set of set of observed variables $X, Y(0), D$ by fitting each component of the Bayes factorization. For each continuous (or discrete) attribute, a linear-Gaussian (or logistic regression) model is fit with respect to its parents. These models are first fit for the baseline time-step ($t=0$) in patient trajectories for the purpose of i) generating the first time step further downstream in the generation process and ii) data imputation for missing values as described in  appendix~\ref{apd:Imputation}. %The marginal root nodes are sampled from a distribution inferred using the statistics observed in the data.

The remaining sequence is fit as an autoregressive model; for each covariate at time $t$, we assume that i) its value is dependent only on its parents in the causal graph at the time $t$ as well as its previous value in the trajectory at time $t-1$. ii) the autoregression is stationary in time. 
%With these assumptions, we again fit linear/logistic models for each covariate given its parents to each observed transition in the data. 
With these models fit, the hand-crafted models for $Z, A, Y(A)$ and tunable parameters  $\{N, \gamma, \epsilon, T, \mu_B\}$, we can now generate $N$ patient trajectories of $T$ time steps with all variables $(Z, X, Y(0), A, Y(A), D)$ through ancestral sampling.
A summary of the fit quality is provided in Appendix~\ref{app:fit}, and cohort statistics in Appendix~\ref{app:stats}.

%
% Benchmark
%
\section{Using the benchmark}
\label{sec:benchmark}
\begin{figure}%[htbp]
\centering
    \subfigure[Number of patient trajectories, $N$]{
    \includegraphics[width=.4\textwidth,valign=t]{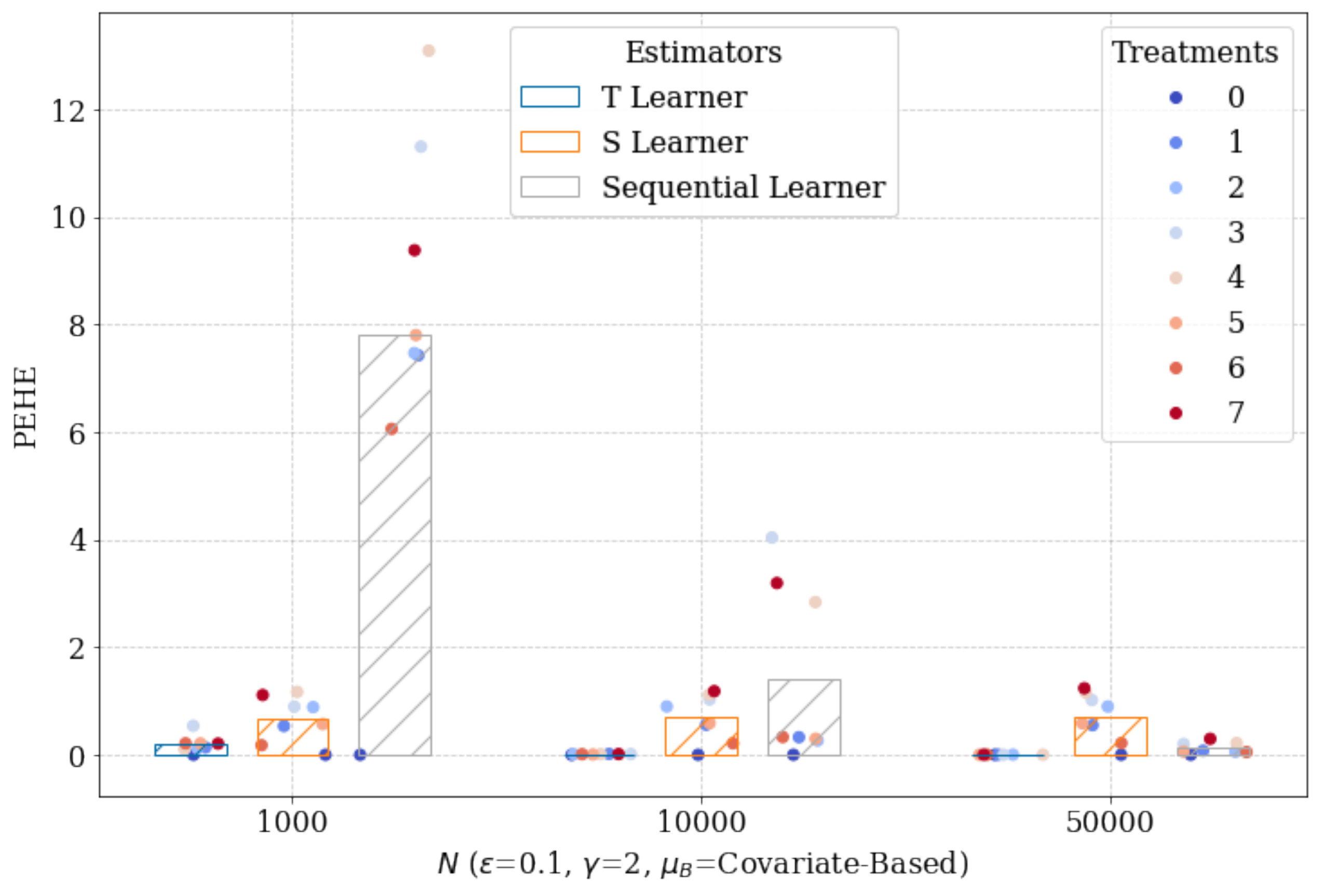}}
    \qquad
    \subfigure[Treatment effect heterogeneity, $\gamma$]{
    \includegraphics[width=.4\textwidth,valign=t]{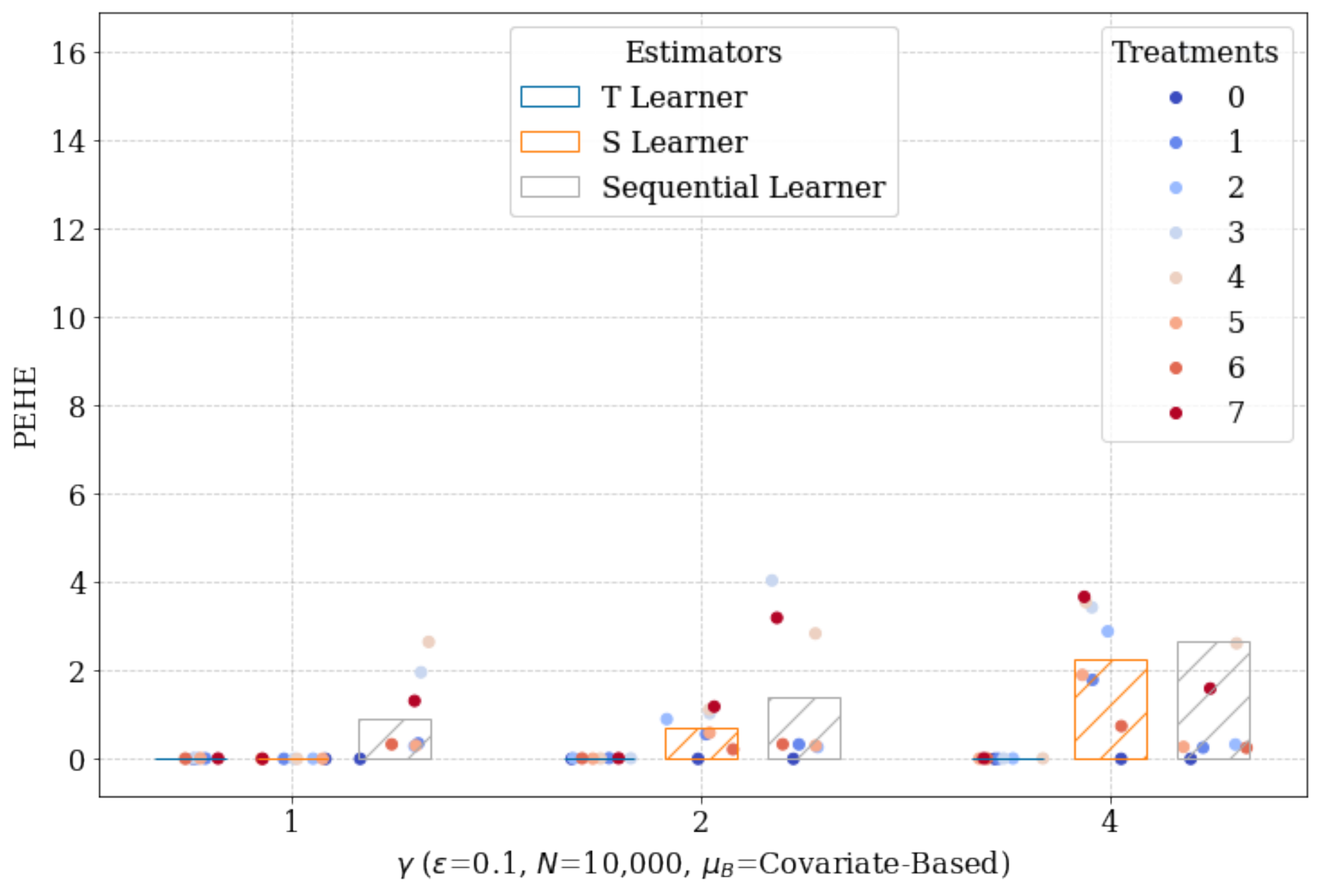}}
    \qquad
    \subfigure[Overlap strength, $\epsilon$]{
    \includegraphics[width=.4\textwidth,valign=t]{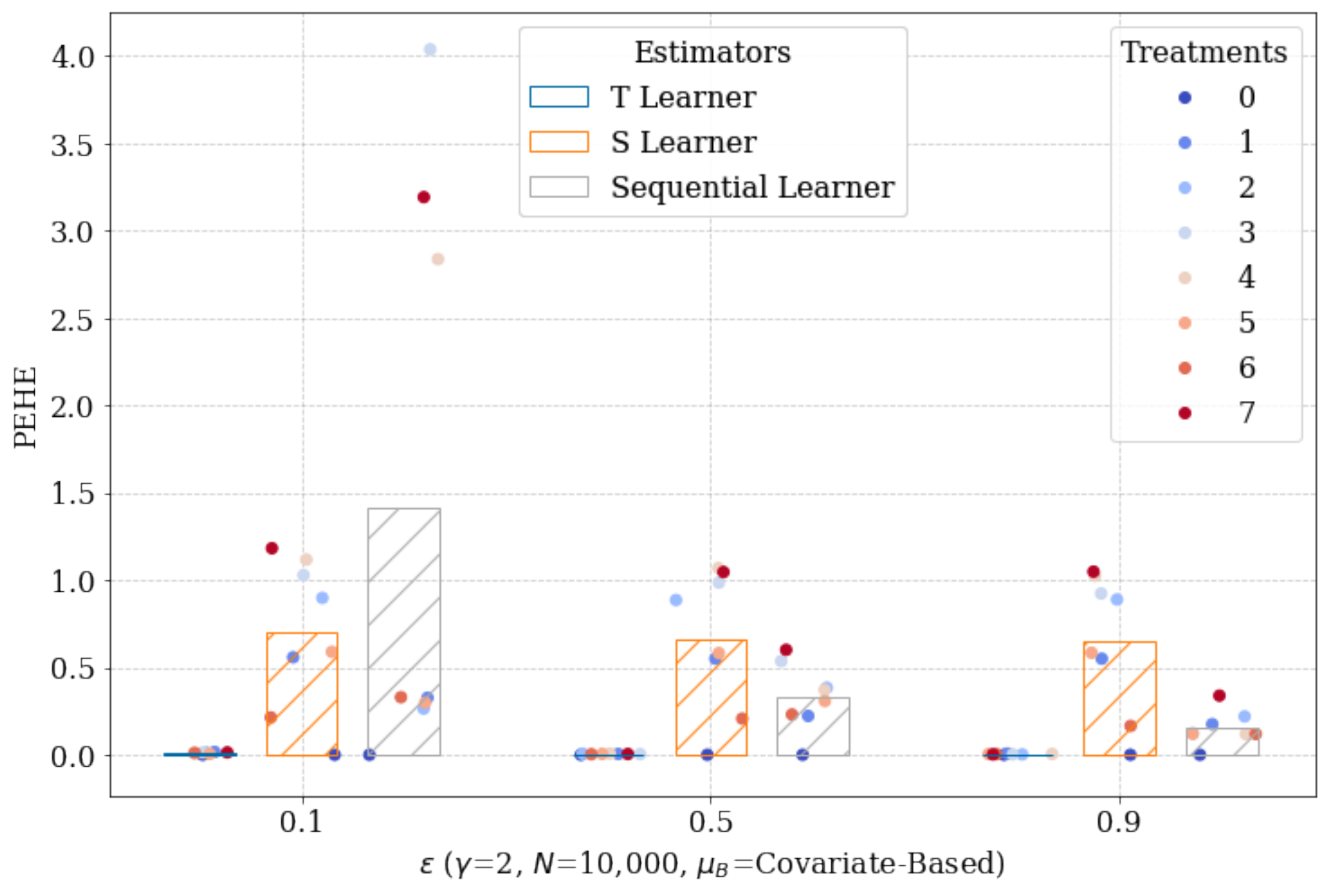}}
    \qquad
    \subfigure[Behavior policy, $\mu_B$]{
    \includegraphics[width=.4\textwidth,valign=t]{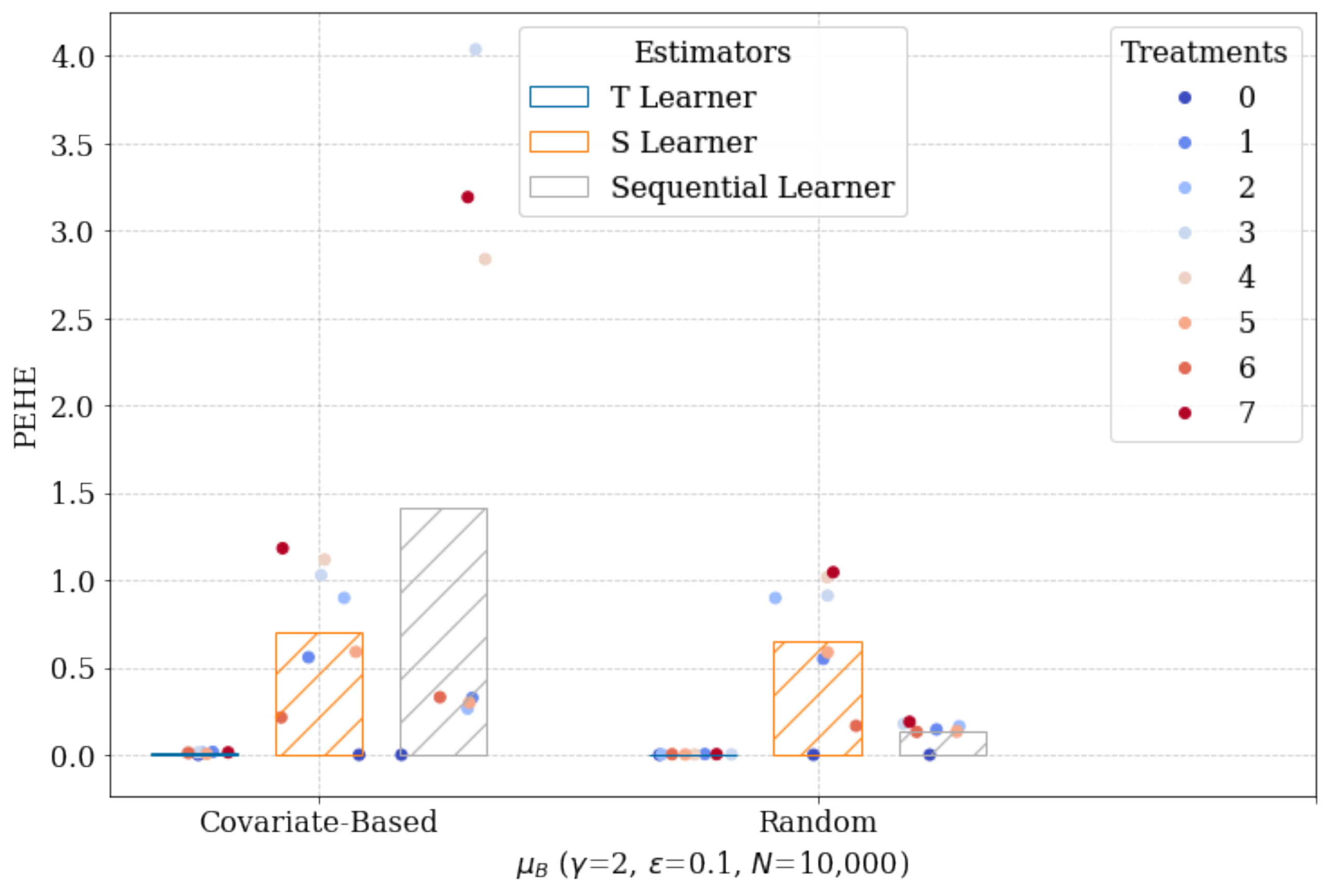}}
    \qquad
    \caption{Comparing the PEHE across different estimators with generated data from varying tunable parameters(N, $\epsilon$, $\gamma$, $\mu_B$) on the ADCB benchmark. }
    \label{fig:knob_results}
\end{figure}

We run experiments aimed at exploring the utility of the generated sequential trajectories and the additional informative covariate sets augmented to the data. The experiments involve the estimation of Average Treatment Effects (ATEs) and Conditional Average Treatment Effects (CATEs). We run them in settings with decisions with single context and in settings where we have a $T$-length sequential context, and compare the precision of estimating heterogeneous effects (PEHE)~\citep{hill2011bayesian} across several estimators whose results are presented in Figure~\ref{fig:knob_results}. 

The estimators presented are S- and T- Learners~\citep{kunzel2019metalearners} with linear regression base learners, as well as a Sequential T-Learner with an RNN base learner to enable incorporation of history. All estimators have access to the disease subtype $Z$ and are compared at the same time step $t=t_s$ in the patient trajectory, with S- and T- Learners trained single-step and the sequential T-Learner trained using sequences \{$t=0, ..., t=t_s$\}.

The results are consistent with expectation for $\epsilon$ and $\mu_B$ due to selection bias and variance due to small $N$. $\gamma$ scales the outcome, so higher $\gamma$ potentially increases the variance, and PEHE becomes harder to predict---heterogeneity is determined by the  variable $Z$, here observed by the estimators. Because linear S-learners cannot capture effect heterogeneity, the  T-learner achieves a lower PEHE. 

%
% Conclusion
%
\section{Conclusion}
\label{sec:conclusion}
We have introduced the Alzheimer's Disease Causal estimation Benchmark (ADCB), a simulation-based platform for evaluating estimators of causal effects and decision-making policies. The simulator is fit to covariates and outcomes from the ADNI database and uses models of treatments and treatment effects inspired by the Alzheimer's disease literature. Usage scenarios for evaluating estimators of causal effects have been presented for varying configurations. Since ADCB generates longitudinal samples of all variables in the system, it can function as a generator of arbitrarily large observational (batch) data, as an online policy learning environment and for design and evaluation of causally adaptive treatment policies. 
More complex confounding models based on the AD literature will be  explored in future iterations of the simulator, increasing the difficulty of the benchmark.

\bibliography{jmlr-sample}

\appendix
\clearpage

\section{Related work}
\label{app:related}
Benchmarking effect estimators is a central problem in causal inference~\citep{wendling2018comparing,shimoni2018benchmarking,gentzel2019case,dorie2019automated,gentzel2021and}. The possibility of producing confounded evaluation metrics prevents using only observational data for this task, without relying on strong assumptions. There are two main approaches which do not rely on such assumptions: a) making use of data from randomized experiments, and b) simulating all or part the system under investigation. See~\citet{gentzel2019case} for an excellent discussion of the pros and cons of each approach. Making use of as much real data for simulation as possible is also called the Empirical Monte Carlo Study (EMCS) approach~\citep{huber2013performance,lechner2013sensitivity}.

Data from randomized experiments are free from confounding bias, which allows for unbiased estimation of causal effects under the (often plausible) assumption of consistency~\citep{gentzel2021and}. To construct a non-trivial benchmark from such data, confounding bias is typically introduced by adding or subsampling data in ways which depend on the contexts of interventions~\citep{hill2011bayesian,kallus2018removing,zhang2021bounding}. The IHDP~\citep{hill2011bayesian} dataset introduces confounding by removing a biased subset of subjects. The famous LaLonde randomized experiment~\citep{lalonde1986evaluating} has been used as a benchmark after mixing in observational (potentially selection-biased) samples~\citep{smith2005does,shalit2017estimating} in the data accessed by estimators. Only the experimental portion is then used for evaluation. Yet another approach is to sample new treatment assignments for study subjects, keeping only the subset of them which agree with the observations from the experiment~\citep{louizos2017causal}. 

A great benefit of using experimental data for evaluation is that it only requires simulating a new treatment assignment policy if the goal is to estimate average treatment effects or the average value of treatment policies~\citep{gentzel2021and}. This is often preferred over simulating the outcome, since treatment policies are within human control, but outcome mechanisms typically are not. Standard or weighted Monte-Carlo estimates are sufficient for evaluation in this case. Drawbacks of this approach include that individual-level counterfactuals and effects remain unknown, and that the sample size is limited to the number of experiment subjects.

If the goal of a benchmark is to evaluate individual-level or fine conditional treatment effects, access to counterfactual outcomes is required. The only way to reliably achieve this is to simulate the mechanism determining the outcome of interventions, which can be done in isolation or in addition to simulating the treatment assignment, as in the Causal Inference Benchmarking Framework by \citet{shimoni2018benchmarking}, the Medkit-Learning environment (focused on reinforcement learning)~\citep{chan2021medkit}, and in IHDP~\citep{hill2011bayesian}. Since the outcome mechanisms are often the main target of estimation, these simulations should be as realistic as possible for the domain they aim to represent. To this end, researchers have considered building their simulators on models fit to observational data~\citep{neal2020realcause,chan2021medkit}.

A drawback of simulated data is that, in many cases, simulators ``tend to match the assumptions of the researcher''~\citep{gentzel2019case}. This is especially problematic in cases where they are introduced to evaluate one particular estimator which may also match those assumptions. As a result, it is important that simulator-based benchmarks contain settings that break or tweak assumptions to appropriately test the robustness of estimators to these. 

\section{Empirical distribution of the A$\beta$ ratio}
\label{apd:abeta}
The the ratio of ($\frac{A\beta-42}{A\beta-40}$) in subjects showing Amyloid-$\beta$ (A$\beta$) plaques form a clearly bimodal distribution;
\begin{figure}[htbp]
    \centering
    \includegraphics[width=0.49\textwidth]{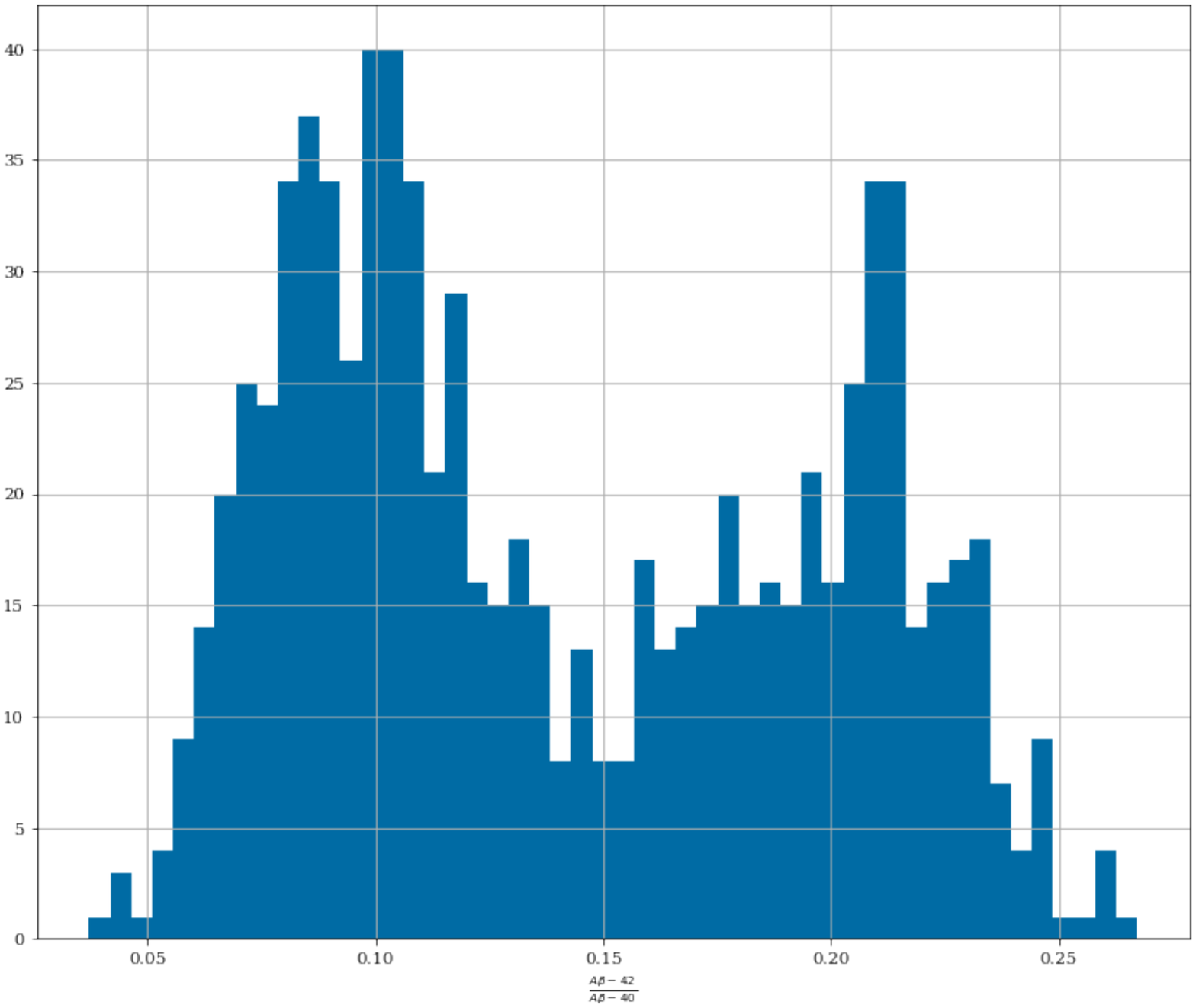}
    \caption{Empirical distribution of the A$\beta$ ratio, used to infer latent disease subtype at baseline.}
    \label{fig:abeta}
\end{figure}

\section{Imputation of missing data}
\label{apd:Imputation}
The patient trajectories have significant missingness along the observation intervals. To cater for the missingness, we impute the missing values based using the causal graph. Our imputation method is inspired by the method Multivariate Imputation by Chained Equations(MICE)~\citep{van2011mice} method but the chaining is done with respect to a variable's parents in the causal graph. For each attribute with a missing value along the time trajectory, we use the linear model learned at baseline to impute the value for that particular attribute at a given timestep using the attribute graph at the timestep of interest.

\section{Temporal dependence between variables in the simulator}
\label{apd:temporal}
The longitudinal model at $t>0$.

\begin{figure}[htbp]
    \centering
    \includegraphics[height=0.205\textheight]{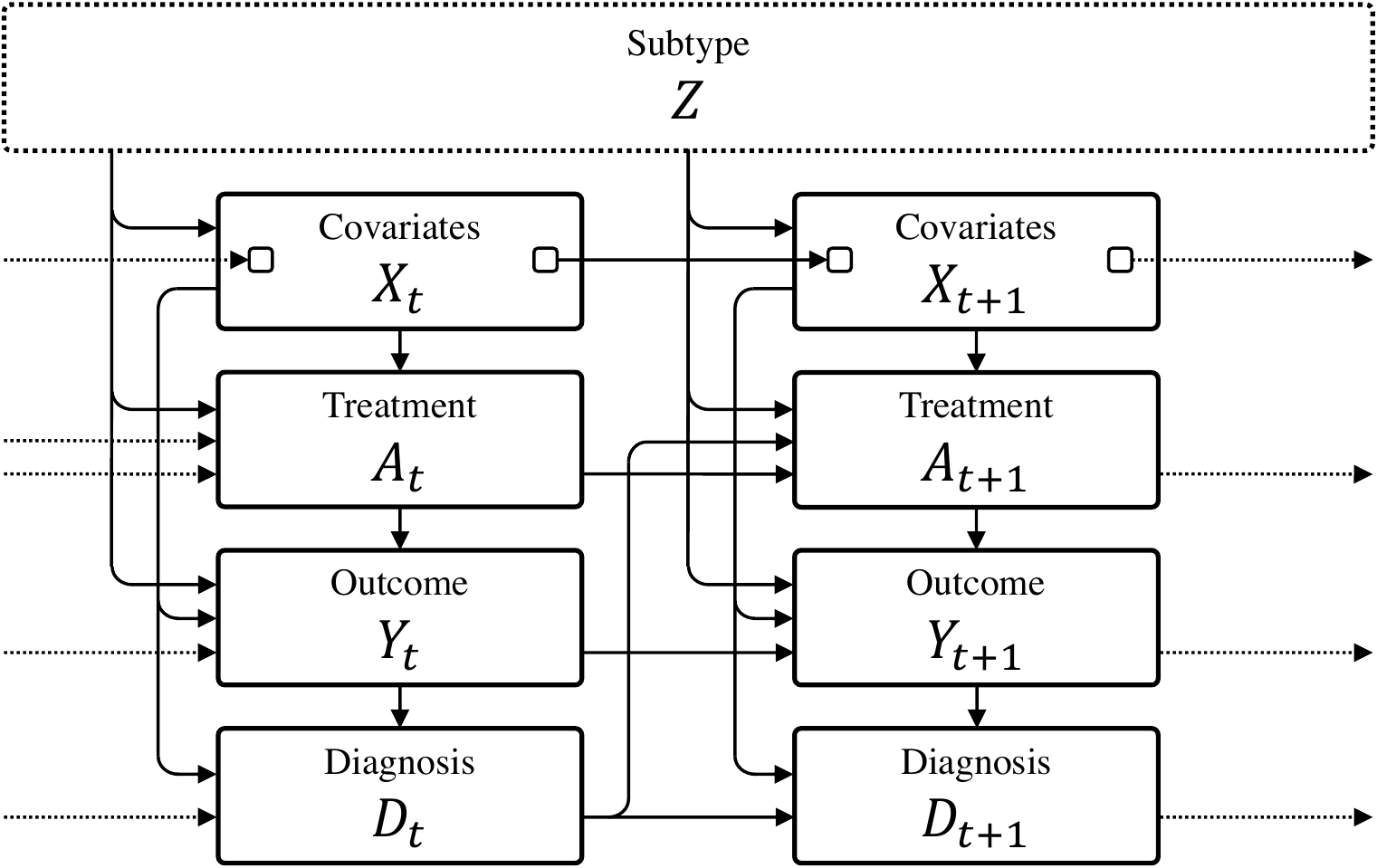}
    \caption{Temporal dependence between variables in the simulator. Each variable obeys the causal dependencies of Figure~\ref{fig:causal_graph} in addition to depending on the previous value of itself. The small box in the set of covariates $X$ indicates that each variable in the set depends only on the previous value of that specific variable. For example, Tau at time $t+1$ depends only on APOE and Race at time $t+1$, the subtype $Z$, and Tau at time $t$. The subtype $Z$ is assumed stationary.}
    \label{fig:causal_graph_time}
\end{figure}

\section{Patient covariates description}
\label{apd:covariates}
The subset of covariates used in this work includes the following and their descriptions as outlined in~\citep{marinescu2018tadpole}  

\begin{enumerate}
    \item FDG PET ROI averages - measure cell metabolism, where cells affected by AD show reduced metabolism
    \item AV45 PET ROI averages - measure amyloid-beta load in the brain, where amyloid-beta is a protein that mis-folds (i.e. its 3D structure is not properly constructed), which then leads to AD
    \item CSF biomarkers - amyloid and TAU levels in the cerebrospinal fluid~(CSF)
    \item Others:
    \begin{itemize}
        \item APOE status - a gene that is a risk factor for developing AD
        \item Demographic information: gender, education, race, marital status ...
        \item Diagnosis: either Cognitively Normal (CN), Mild Cognitive Impairment (MCI) or Alzheimer's disease (AD).
        \end{itemize}
\end{enumerate}

\section{Average Treatment Effects from Literature}
\label{apd:grossberg}
\begin{table}[!hbtp]
\floatconts
  {tab:example-booktabs}
 {\caption{ Average treatment effects (ATE), in terms of change in ADAS-Cog compared to no treatment, of various therapies from meta-analyses of clinical trials ~\citep{grossberg2019present}}}%
  {\begin{tabular}{llc}
  \toprule
  \bfseries {a} & \bfseries {Treatment} & \bfseries {ATE} $\tau(a)$\\
  \midrule
  0 & No treatment & 0 \\
  \midrule
  1 & Donepezil 5 mg & –1.95\\
  2 & Donepezil 10 mg & –2.48 \\
  3 & Galantamine 24 mg & –3.03\\
  4 & Galantamine 32 mg & –3.20\\
  5 & Rivastigmine 12 mg & –2.01\\
  6 & Memantine 20 mg & –1.29\\
  7 & Memantine + ChEI & –2.64\\
  \bottomrule
  \end{tabular}}
\end{table}

\begin{table*}[t]
    \centering
    \caption{\label{tab:model_fit} Baseline and autoregression model performance. }
    \begin{tabular}{l|ccc|ccc}
        \toprule
        Target variable & \multicolumn{3}{c|}{Baseline} & \multicolumn{3}{c}{Autoregression} \\
        \toprule
        Classifiers & Acc & F1 & \# Classes  & Acc & F1 & \# Classes \\
        \midrule
        APOE4 & 52\% & 0.38 & 3 & 100\% & 1.00 & 3 \\
        Education (years) & 21\% & 0.09 & 13 & 100\% & 1.00 & 10 \\
        Marital status & 73\% & 0.62 & 5 & 96\% & 0.94 & 4 \\
        Diagnosis & 70\% & 0.70 & 3 & 84\% & 0.82 & 3 \\
        \toprule
        Regressions & $R^2$ & RMSE & $\sigma_Y$ & $R^2$ & RMSE & $\sigma_Y$ \\
        \midrule
        Tau & -1.92 & 113.15 & 1.3e+02 & 0.83 & 43.69 & 1.2e+02 \\
        PTau & -1.38 & 12.02 & 14 & 0.94 & 2.96 & 14 \\
        FDG & -3.60 & 0.14 & 0.15 & -0.99 & 0.07 & 0.087 \\
        AV45 & 0.25 & 0.15 & 0.24 & -15.40 & 0.12 & 0.12 \\
        ADAS13 & -0.05 & 6.89 & 9.4 & 0.40 & 6.71 & 6.3 \\
%        CDRSB & -0.28 & 1.32 & 1.5 & -0.34 & 2.20 & 2.2 \\
%        MMSE & -0.59 & 2.03 & 2.6 & -0.30 & 1.51 & 1.4 \\
         \bottomrule
    \end{tabular}
\end{table*}

\section{Model fit}
\label{app:fit}
We evaluate model fit independently for each variable, as summarized in Table~\ref{tab:model_fit}. The overall predictability for baseline variables was low in general, with non-trivial accuracy attained only for a handful of the covariates, including diagnosis and AV45 levels. Autoregressors achieved significantly better results due to some variables being more or less static and time or varying very slowly.

\section{Cohort statistics}
\label{app:stats}
Overview statistics for synthtetic and real-world cohorts are presented in Table~\ref{tab:cohorts}. 

\begin{table*}[t]
\centering
\caption{\label{tab:cohorts}Cohort statistics for the first timestep (T=1) for simulated (ADCB) and observed real-world subjects (ADNI). Continuous variables are described by mean (standard deviation) and categorical variables by count (frequency in $\%$).}
\begin{tabular}{lcc}
\toprule  & ADCB T=1, n=10000 & ADNI T=1, n=844\\ 
\midrule 
Demographics\\ 
\midrule 
Gender \\
\ \ Female & 4584 (45.8$\%$) & 395 (46.8$\%$) \\ 
\ \ Male & 5416 (54.2$\%$) & 449 (53.2$\%$) \\ 
Marital status \\
\ \ Divorced & 7651 (76.5$\%$) & 634 (75.1$\%$) \\ 
\ \ Married & 323 (3.2$\%$) & 29 (3.4$\%$) \\ 
\ \ Never married & 1121 (11.2$\%$) & 96 (11.4$\%$) \\ 
\ \ Unknown & 836 (8.4$\%$) & 80 (9.5$\%$) \\ 
\ \ Widowed & 69 (0.7$\%$) & 5 (0.6$\%$) \\ 
Ethnicity \\
\ \ Hisp/Latino & 341 (3.4$\%$) & 30 (3.6$\%$) \\ 
\ \ Not Hisp/Latino & 9603 (96.0$\%$) & 809 (95.9$\%$) \\ 
\ \ Unknown & 56 (0.6$\%$) & 5 (0.6$\%$) \\ 
Race \\
\ \ Am Indian/Alaskan & 9256 (92.6$\%$) & 783 (92.8$\%$) \\ 
\ \ Asian & 371 (3.7$\%$) & 31 (3.7$\%$) \\ 
\ \ Black & 154 (1.5$\%$) & 13 (1.5$\%$) \\ 
\ \ Hawaiian/Other PI & 16 (0.2$\%$) & 1 (0.1$\%$) \\ 
\ \ More than one & 158 (1.6$\%$) & 12 (1.4$\%$) \\ 
\ \ Unknown & 28 (0.3$\%$) & 2 (0.2$\%$) \\ 
\ \ White & 17 (0.2$\%$) & 2 (0.2$\%$) \\ 
Education & 13.2 (2.7) & 13.3 (2.6) \\ 
\midrule 
Biomarkers\\ 
\midrule 
Tau & 285.9 (123.3) & 279.6 (130.0) \\ 
PTau & 27.3 (13.0) & 26.7 (14.2) \\ 
FDG & 1.2 (0.2) & 1.2 (0.2) \\ 
AV45 & 1.2 (0.3) & 1.2 (0.2) \\ 
APOE4 \\
\ \ 0.0 & 5564 (55.6$\%$) & 460 (54.5$\%$) \\ 
\ \ 1.0 & 3525 (35.2$\%$) & 303 (35.9$\%$) \\ 
\ \ 2.0 & 911 (9.1$\%$) & 81 (9.6$\%$) \\ 
\midrule 
Outcomes\\ 
\midrule 
ADAS13 & 16.1 (8.3) & 15.4 (9.5) \\ 
Diagnosis \\
\ \ CN & 3392 (33.9$\%$) & 275 (32.6$\%$) \\ 
\ \ Dementia & 5026 (50.3$\%$) & 438 (51.9$\%$) \\ 
\ \ MCI & 1582 (15.8$\%$) & 131 (15.5$\%$) \\ 
\midrule 
\iffalse
Treatments\\ 
\midrule 
Treatment \\
\ \ No treatment & 10000 (100.0$\%$) & 844 (100.0$\%$) \\ 
\ \ Donepezil 5mg & 0 (0.0$\%$) & 0 (0.0$\%$) \\ 
\ \ Donepezil 10mg & 0 (0.0$\%$) & 0 (0.0$\%$) \\ 
\ \ Galantamine 24mg & 0 (0.0$\%$) & 0 (0.0$\%$) \\ 
\ \ Galantamine 32mg & 0 (0.0$\%$) & 0 (0.0$\%$) \\ 
\ \ Rivastigmine 12mg & 0 (0.0$\%$) & 0 (0.0$\%$) \\ 
\ \ Memantine 20mg & 0 (0.0$\%$) & 0 (0.0$\%$) \\ 
\ \ Memantine+ChEI & 0 (0.0$\%$) & 0 (0.0$\%$) \\ 
\midrule 
\fi
Subtype\\ 
\midrule 
Subtype & 4179 (41.8$\%$) & - (-) \\ 
\bottomrule 
\end{tabular}
\end{table*}

\end{document}